\documentclass[journal,twocolumn,final]{IEEEtran}

\usepackage[utf8]{inputenc} 
\usepackage[T1]{fontenc}    
\usepackage{url}            
\usepackage[colorlinks]{hyperref}       
\usepackage{amsfonts}       
\usepackage[dvipsnames]{xcolor}         
\usepackage{amsmath}
\usepackage{amssymb}
\usepackage{graphicx}
\usepackage{subcaption}
\usepackage{float}
\usepackage{algorithm}
\usepackage{algorithmic}

\usepackage{tikz}
\usetikzlibrary{positioning}
\usetikzlibrary{decorations.pathmorphing,patterns,snakes}
\usepackage{svg}

\usepackage{balance}

\PassOptionsToPackage{hyphens}{url}
\hypersetup{
    linkcolor=blue,
    filecolor=magenta,      
    urlcolor=teal,
    citecolor=magenta
    }

\title{Beyond Convergence: Identifiability of Machine Learning and Deep Learning Models}

\author{Reza~Sameni\textsuperscript{*}
\thanks{R.~Sameni is with the Department of Biomedical Informatics, Emory University, Atlanta, GA, USA. Email: \href{mailto:rsameni@dbmi.emory.edu}{rsameni@dbmi.emory.edu}\\
Supplementary material for the ``BMI-532: Model-based machine learning'' course presented at Emory University, Spring 2023.
}}

\begin{document}
\maketitle
\begin{abstract}
Machine learning (ML) and deep learning models are extensively used for parameter optimization and regression problems. However, not all inverse problems in ML are ``identifiable,'' indicating that model parameters may not be uniquely determined from the available data and the data model's input-output relationship.

In this study, we investigate the notion of model parameter identifiability through a case study focused on parameter estimation from motion sensor data. Utilizing a bipedal-spring mass human walk dynamics model, we generate synthetic data representing diverse gait patterns and conditions. Employing a deep neural network, we attempt to estimate subject-wise parameters, including mass, stiffness, and equilibrium leg length. The results show that while certain parameters can be identified from the observation data, others remain unidentifiable, highlighting that unidentifiability is an intrinsic limitation of the experimental setup, necessitating a change in data collection and experimental scenarios.

Beyond this specific case study, the concept of identifiability has broader implications in ML and deep learning. Addressing unidentifiability requires proven identifiable models (with theoretical support), multimodal data fusion techniques, and advancements in model-based machine learning. Understanding and resolving unidentifiability challenges will lead to more reliable and accurate applications across diverse domains, transcending mere model convergence and enhancing the reliability of machine learning models.
\end{abstract}


\section{Introduction}
Machine learning (ML) and deep learning models have demonstrated remarkable capacity to train and optimize model parameters based on data. The quality of adaptation in these models is commonly assessed using standard convergence measures such as mean squared error (MSE), mean absolute error (MAE), accuracy, etc. However, even with abundant training data and successful convergence, not all inverse problems are ``identifiable,'' in the sense that the model parameters may be uniquely determined from the available data.

Identifiability, a fundamental concept in system and control theory \cite{audoly2001global} and stochastic inference \cite{pearl2009causality}, rigorously examines the possibility of uniquely determining model parameters based on the (mathematical) description of the input-output relationship. Certain models may remain unidentifiable regardless of data volume, quality, and model accuracy due to the data-parameter relationship. Unidentifiable problems appear at the heart of many data modeling and stochastic inference problems, including retrieving time-series data from noisy measurements, estimating likelihoods or log-likelihoods in Bayesian inference problems, or training the weights of shallow and deep neural networks \cite{Roeder2020,Kivva2022}. Although unidentifiability is an intrinsic property of the data model that relates the measurements and parameters, it may go unnoticed, particularly in complex ML and deep learning models with thousands or millions of hidden parameters. This often occurs when models exhibit successful convergence during training or when the framework is fully data-driven without explicit data models. Despite the apparent training convergence, unidentifiable models can pose potential misinterpretation, especially when the underlying data models are not analytically available or easy to study.

In this work, we study the notion of identifiability through a case study focused on human gait and motion analysis. We employ the well-known bipedal-spring mass human walk dynamics model developed by Geyer et al \cite{geyer2006compliant}. Our objective is to identify subject-wise parameters, such as weight, height, and leg stiffness, by analyzing a person's walking data, with potential applications in identifying pathological cases. We will use Geyer's model as the base model for generating synthetic human motion data. The synthetic data will then be used to train deep learning models to estimate subject-wise parameters. Our case study will highlight the intrinsic and recurring challenge of identifiability in machine learning, an aspect that has received limited attention in the literature.

While we do not provide a definitive solution for unidentifiable ML problems, we shed light on this relatively unexplored aspect, which is crucial for ensuring the generalizability and explainability of state-of-the-art machine and deep learning models. By underscoring this recurrent issue, we emphasize the necessity of considering identifiability during model development and analysis, ultimately leading to more informed models across diverse domains. Furthermore, we demonstrate the potential of \textit{model-based machine learning} to identify and potentially diagnose unidentifiable models.

The paper is organized as follows: In Section \ref{sec:data_model}, we present Geyer's human gait model, which we will use for generating synthetic data and explore the identifiable and unidentifiable parameters of this model. In Section \ref{sec:model_development} we present the deep learning architecture, followed by the training, validation and test results in Section~\ref{sec:results}, to estimate subject-wise motion parameters. The paper concludes with a discussion on potential ideas for studying model identifiability in general machine learning and deep learning models, offering insights into the identifiability challenges in the context of model-based machine learning and deep learning.

\section{Data Model}
\label{sec:data_model}
\begin{figure*}[tb]
  \centering
  \begin{tikzpicture}[remember picture]
    %
    \draw[thick,gray,line width=0.75mm] (0,1.4) -- (13,1.4);
    \draw [-stealth,line width=0.3mm](0,1.4) -- (.5,1.4);
    \draw [-stealth,line width=0.3mm](0,1.4) -- (0,1.9);
    \draw[dotted,line width=0.3mm] (2.4, 1.8) -- (3.0,1.8);
    \draw[dotted,line width=0.3mm] (11.2, 1.8) -- (11.7,1.8);
    \draw[dotted, black, line width=0.4mm, smooth] plot coordinates{(0.8,3.2) (1.95,3.5) (4.95,3.0) (6.5,2.85) (8.0,3.0) (10.7,3.5) (12,3.2)};   
    \draw[-,snake=zigzag, line width=0.25mm, segment amplitude=.6mm, segment length=1.5mm, line before snake=6mm, line after snake=5.0mm, line width= 0.5mm] (1.95,3.5) -- (1.95,1.5);
    \draw [fill] (1.95,1.5) circle [radius=2pt];
    \draw[-,snake=zigzag, line width=0.25mm, segment amplitude=.6mm, segment length=1.5mm, line before snake=6mm, line after snake=5.0mm, line width= 0.5mm, gray] (1.95,3.5) -- (3.0,1.8);
    \draw [fill=gray] (3.0,1.8) circle [radius=2pt];
    \draw [fill=brown] (1.95,3.5) circle [radius=8pt];
    \draw[-,snake=zigzag, line width=0.25mm, segment amplitude=.6mm, segment length=1.5mm, line before snake=6mm, line after snake=5.0mm, line width= 0.5mm] (4.95,3.0) -- (4.5,1.5);
    \draw [fill] (4.5,1.5) circle [radius=2pt];
    \draw[-,snake=zigzag, line width=0.25mm, segment amplitude=.6mm, segment length=1.5mm, line before snake=6mm, line after snake=5.0mm, line width= 0.5mm, gray] (4.95,3.0) -- (5.95,1.5);
    \draw [fill=gray] (5.95,1.5) circle [radius=2pt];
    \draw [fill=brown] (4.95,3.0) circle [radius=8pt];
    \draw[-,snake=zigzag, line width=0.25mm, segment amplitude=.6mm, segment length=1.5mm, line before snake=6mm, line after snake=5.0mm, line width= 0.5mm] (8.0,3.0) -- (7.1,1.5);
    \draw [fill] (7.1,1.5) circle [radius=2pt];
    \draw[-,snake=zigzag, line width=0.25mm, segment amplitude=.6mm, segment length=1.5mm, line before snake=6mm, line after snake=5.0mm, line width= 0.5mm, gray] (8.0,3.0) -- (8.5,1.5);
    \draw [fill=gray] (8.5,1.5) circle [radius=2pt];
    \draw [fill=brown] (8.0,3.0) circle [radius=8pt];
    \draw[-,snake=zigzag, line width=0.25mm, segment amplitude=.6mm, segment length=1.5mm, line before snake=6mm, line after snake=5.0mm, line width= 0.5mm,gray] (10.7,3.5) -- (10.7,1.5);
    \draw [fill=gray] (10.7,1.5) circle [radius=2pt];
    \draw[-,snake=zigzag, line width=0.25mm, segment amplitude=.6mm, segment length=1.5mm, line before snake=6mm, line after snake=5.0mm, line width= 0.5mm] (10.7,3.5) -- (11.7,1.8);
    \draw [fill] (11.7,1.8) circle [radius=2pt];
    \draw [fill=brown] (10.7,3.5) circle [radius=8pt];   %
    \node[anchor=west] at (0.2, 1.1) {$x$};
    \node[anchor=east] at (0, 1.7) {$y$};
    \node[anchor=west] at (1.4, 2.5) {$k$};
    \node[anchor=west] at (2.5, 2.8) {$\ell_0$};
    \node[anchor=west] at (0.5, 4.2) {$\textsl{g}$};     \draw [-stealth,line width=0.4mm](0.5,4.4) -- (0.5,3.8);
    \node[anchor=west] at (7.6, 3.5) {$m$};
    \node[anchor=west] at (2.2, 2.0) {$\alpha_0$};
    \node[anchor=west] at (11, 2.0) {$\alpha_0$};
    \node[anchor=west] (node1) at (1.1, 4.2) {apex $i$};
    \node[anchor=west] (node2) at (10.2, 4.2) {apex $i+1$};
    \draw[->, >=stealth, line width=1pt,gray] (node1) -- (node2) node[midway, above] {single step};
    \node[anchor=west] at (1.65, 1.1) {$\text{FP}_i$};
    \node[anchor=west] at (10.4, 1.1) {$\text{FP}_{i+1}$};
    \node[anchor=west] at (0.6, .3) {left single support};
     \draw[|-|] (0.5,.8) -- (3.5,.8);
    \node[anchor=west] at (5.5, .3) {double support};
     \draw[|-|] (3.7,.8) -- (9.2,.8);
    \node[anchor=west] at (9.4, .3) {right single support};
     \draw[|-|] (9.4,.8) -- (12.2,.8);
  \end{tikzpicture}
  \caption{Compliant leg behavior explains basic dynamics of walking and running; adapted from (Geyer et al. 2006) \cite{geyer2006compliant}.}
  \label{fig:Geyer2006_Model}
\end{figure*}
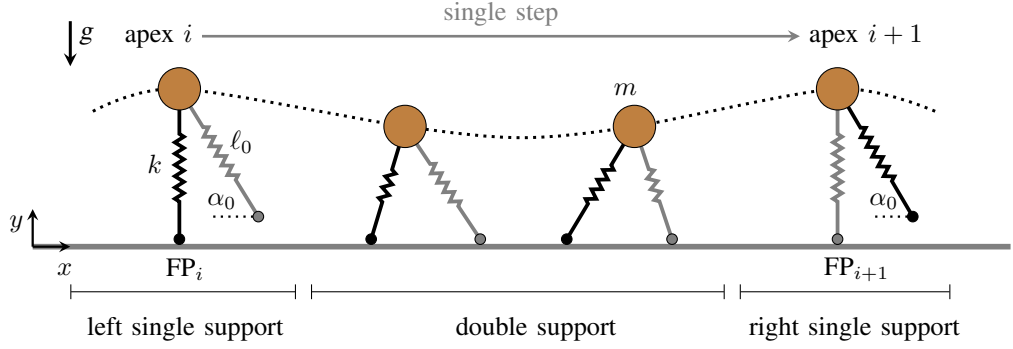

The model proposed by Geyer et al. provides a simplified physics-based representation of human walking and running dynamics \cite{geyer2006compliant}. The model describes the fundamental dynamics of human walking and running through the concept of compliant leg behavior. It incorporates the interaction between the center of mass and the compliant legs during different phases of human gait, including single support, double support, and the transition between steps, as shown in Fig.~\ref{fig:Geyer2006_Model}. This compliant leg behavior, characterized by the mechanical properties of the legs to absorb and release energy during the stance phase of gait, plays a crucial role in accurately capturing the observed dynamics of human locomotion.


According to Fig.~\ref{fig:Geyer2006_Model}, the center of mass dynamics of the bipedal-spring mass human walking model can be described as follows for a single step from apex $i$ to $i+1$: (i) left leg single support: $m\ddot{x} = P x$, $m\ddot{y} = Py - m g$; (ii) intermittent double support: $m\ddot{x} = P x - Q (d - x)$, $m\ddot{y} = Py + Q y- m g$; (iii) right leg single support: $m\ddot{x} = - Q (d - x)$, $m\ddot{y} = Q y- m g$, where $m$ is the total mass of the human body; $k$ is the stiffness parameter representing the compliant behavior of the legs; $\ell_0$ is the equilibrium leg length, denoting the length of the legs at rest; $\textsl{g}$ is the gravitational acceleration; $x$ and $y$ denote the horizontal and vertical positions of the center of mass, respectively. Additionally, the model employs two stiffness parameters to represent the compliant behavior of the legs during different phases of gait: $P = k (\ell_0/\sqrt{x^2 + y^2} - 1)$ for single support phases and $Q = k (\ell_0\sqrt{(d - x)^2 + y^2} - 1)$ for double support phases, where $d = \text{FP}_{i+1,x} - \text{FP}_{i,x}$ represents the horizontal distance covered in each step. The foot point (FP) refers to the point on the ground where the leg stance springs attach during gait. These equations collectively define the dynamics of the bipedal-spring mass human walk model \cite{geyer2006compliant}.

In the sequel, we will use Geyer's model as a forward model to generate synthetic human walk data with different subject-wise parameters $(m, k, \ell_0)$. The research question is whether we can fit this model onto real data and train a neural network to estimate the model parameters $m$, $k$, and $\ell_0$ from real measurements. The application of such a model includes extracting physics-based features from motion sensors attached to patients, allowing for the study and classification of their walking patterns for normal and pathological cases such as tremulous patients.

\subsection{Identifiability of Geyer's model}
Before training a machine/deep learning model, let us revisit Geyer's data model from an identifiability perspective. Normalizing both sides of the the bipedal-spring mass equations by the mass $m$, the governing equations that describe the three phases of motion in Fig.~\ref{fig:Geyer2006_Model} can be rewitten as follows:

\begin{itemize}
    \item \textbf{Left leg single support:} 
    $\ddot{x} = p x$, $\ddot{y} = py - g$
  
    \item \textbf{Intermittent double support:} 
    $\ddot{x} = p x - q (d - x)$, $\ddot{y} = py + q y- g$
    
    \item \textbf{Right leg single support:} 
    $\ddot{x} = - q (d - x)$, $\ddot{y} = q y- g$
\end{itemize}
where $p = \rho (\ell_0/\sqrt{x^2 + y^2} - 1)$, $q = \rho (\ell_0/\sqrt{(d - x)^2 + y^2} - 1)$ and the new variable $\rho=k/m$ is the stiffness to mass ratio. Apparently, in this model, only the ratio of $m/k$ impacts the motion dynamics and the observed $(x, y)$, and not $m$ and $k$ alone. This suggests that from the measurements $(x,y)$, $m$ and $k$ are \textit{mathematically unidentifiable}, and no estimation framework will be able to estimate both parameters independently. However, $\ell_0$ and the ratio $\rho$ may still be identifiable. Apparently, this is an intrinsic limitation of the physical data model, which may not be resolved by increasing the size of training data or even the parameter estimation scheme. In the sequel, we will assess this fact from a data-drive perspective.

\section{A deep learning model for estimating gait parameters from motion sensor data}
\label{sec:model_development}
We are interested in designing a deep learning model to estimate subject-wise gait parameters $m$, $k$ and $\ell_0$, from motion sensor data collected during human walking.

To assess the feasibility of this goal, we utilize the Geyer's model detailed in Section~\ref{sec:data_model}, to generate synthetic data and will use the data to train a neural network for subject-wise parameter estimation. This synthetic data generation process enables us to build a robust dataset representing a wide range of gait patterns and conditions. The trained model can eventually used to identify subject-wise parameters from real data.

To create diverse synthetic data, we used Geyer's model to generate 1,000,000 records, each corresponding to 15\,s of data with a time step of 0.1\,s. The gravitational acceleration was fixed to $\textsl{g} = 10$\,m/s\textsuperscript{2}. The subject-wise parameters were selected from random distributions: $m \sim U[60, 75]$\,kg, stiffness $k \sim U[8000, 10000]$\,N/m, equilibrium leg length $\ell_0 \sim U[0.6, 0.8]$\,m, initial horizontal position $x_0 \sim U[0, 0.1]$\,m, and initial vertical position $y_0 \sim U[0, 0.1]$\,m, where $U[a, b]$ denotes uniform distribution between $a$ and $b$.

\begin{figure}[tb]
    \centering
    \includegraphics[width=0.6\columnwidth]{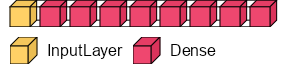}
    \caption{Overview of the neural network architecture for parameter estimation. The output layer has three outputs $(m,k,\ell_0)$ or two outputs $(\rho,\ell_0)$, corresponding to the two studied scenarios described in Section~\ref{sec:model_development}.}
\label{fig:dnn_architecture}    
\end{figure}

The neural network architecture designed for parameter estimation consists of multiple layers, including an input layer, 7 fully connected layers with 50 nodes, and an output layer as shown in Fig.~\ref{fig:dnn_architecture}.  The number of model training epochs was set to 500, with a training batch size of 1000. The 1-million record synthetic dataset was split into 80\% for training, 10\% for validation, and 10\% for testing, ensuring a comprehensive evaluation of the deep learning model's performance.

\begin{figure}[tb]
    \centering
    \begin{subfigure}{0.45\textwidth}
        \centering     \includegraphics[width=0.8\columnwidth]{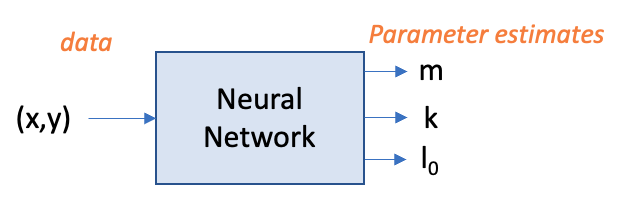}
        \caption{Three parameter model}\label{fig:3_params_model}
    \end{subfigure}
    \hfill
    \begin{subfigure}{0.45\textwidth}
        \centering
        \includegraphics[width=0.8\columnwidth]{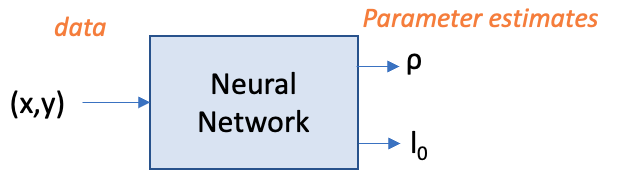}
        \caption{Two parameter model}\label{fig:2_params_model}
    \end{subfigure}
    \caption{Overview of the two model training scenarios, using the subject's center of mass position $(x,y)$ as input and the parameters $(m, k, \ell_0)$ or $(\rho,\ell_0)$ as output.}
    \label{fig:parameter_estimation_modes}
\end{figure}

Two regression scenarios were studied to investigate the notion of identifiability:
\begin{enumerate}
    \item Using $(x,y)$ as inputs to estimate all three parameters: $(m,k,\ell_0)$ in the output (Fig.~\ref{fig:3_params_model}).
    \item Using $(x,y)$ as inputs to estimate $(\ell_0,\rho)$ in the output (Fig.~\ref{fig:2_params_model}).
\end{enumerate}
From Section~\ref{sec:data_model}, we know that the first scenario is unidentifiable; but the second scenario may potentially be identifiable (to be verified by model training on the synthetic data).



\section{Results}
\label{sec:results}
The training and validation results of the two scenarios described in Section~\ref{sec:model_development} for estimating the parameters $(m, k, \ell_0)$ and $(\rho, \ell_0)$ are presented below.
\subsection{The three-parameter regression problem}
\begin{figure}[tb]
    \centering
     \begin{subfigure}[b]{.49\columnwidth}
         \centering
         \includegraphics[trim=0 0in 0in 0in, clip, width=\columnwidth]{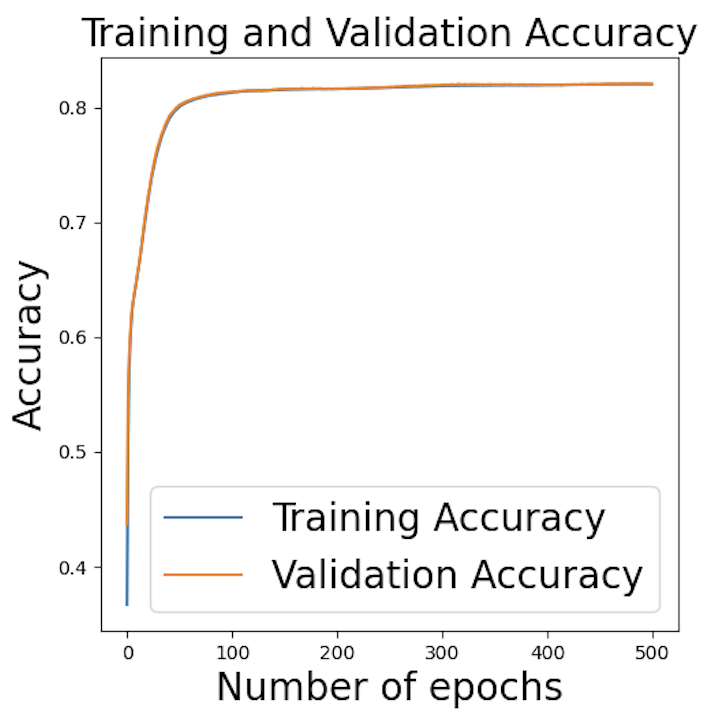}
         \caption{Accuracy}
         \label{fig:accuracy}
     \end{subfigure}
     \begin{subfigure}[b]{.49\columnwidth}
         \centering
         \includegraphics[trim=0in 0in 0 0in, clip, width=\columnwidth]{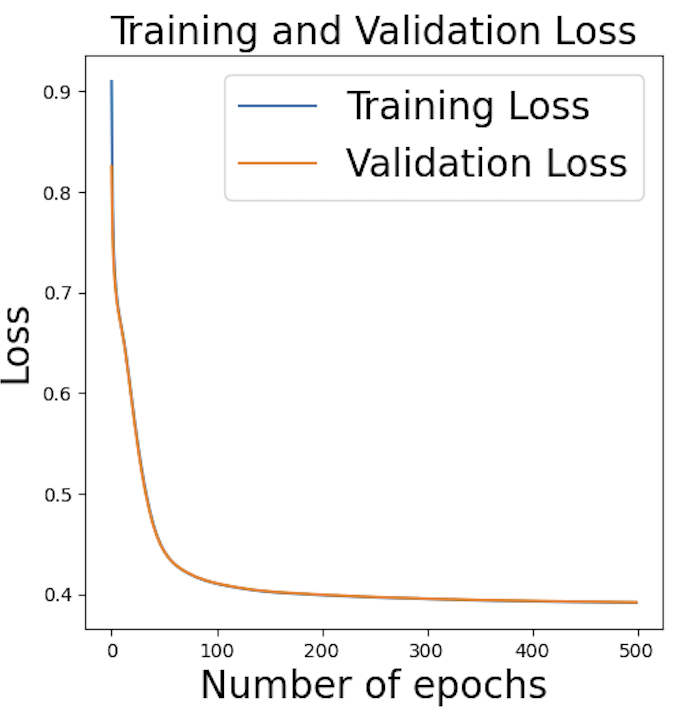}
         \caption{Loss}
         \label{fig:loss}
     \end{subfigure}     
    \caption{Training and validation set accuracy and loss for the three-parameter model: $(m, k, \ell_0)$.}
    \label{fig:training_validation_loss}
\end{figure}

In this case, the deep learning model demonstrated successful convergence and generalization during the training process. Fig.~\ref{fig:training_validation_loss} illustrates the accuracy and loss on the training and validation sets. The decreasing training loss and stable validation loss indicate the model's ability to adapt to the synthetic data and generalize to unseen samples. 

\begin{figure*}[tb]
     \centering
     \begin{subfigure}[b]{0.3\textwidth}
         \centering
         \includegraphics[trim={0in 0 .5in .7in},clip,width=.95\textwidth]{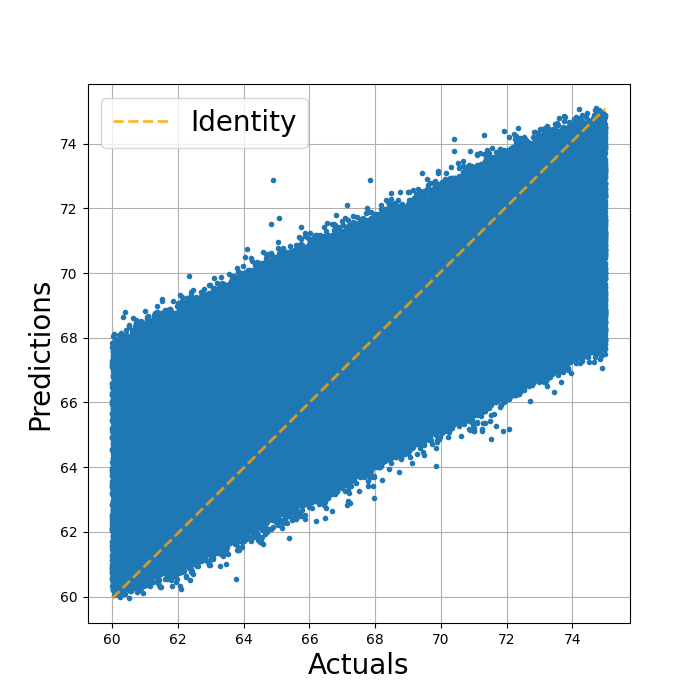}
         \caption{$m$}
         \label{fig:pred_vs_actual_0}
     \end{subfigure}
     \hfill
     \begin{subfigure}[b]{0.3\textwidth}
         \centering
         \includegraphics[trim={0in 0 .5in .7in},clip,width=.95\textwidth]{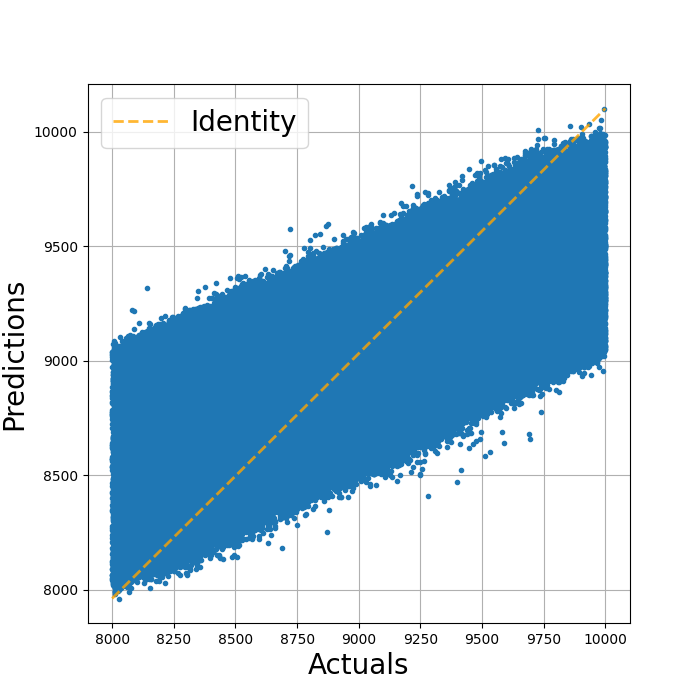}
         \caption{$k$}
         \label{fig:pred_vs_actual_1}
     \end{subfigure}
     \hfill
     \begin{subfigure}[b]{0.3\textwidth}
         \centering
         \includegraphics[trim={0in 0 .5in .7in},clip,width=.95\textwidth]{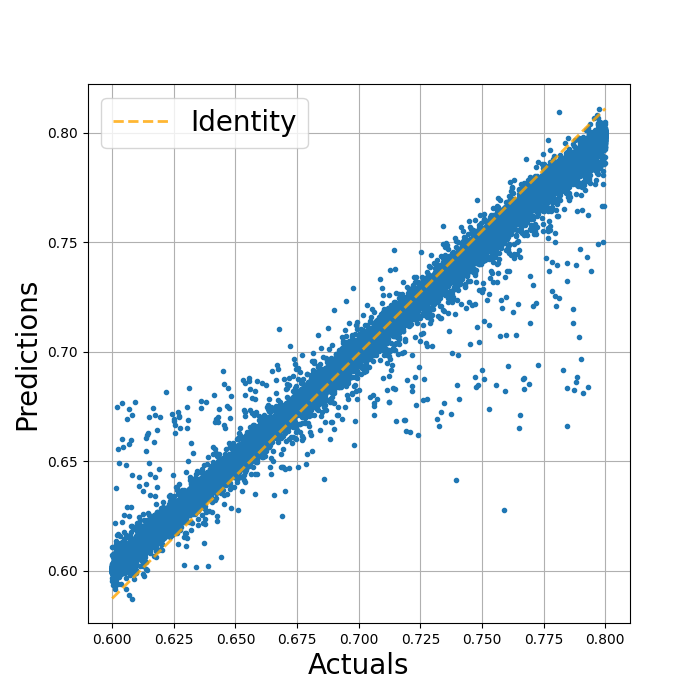}
         \caption{$\ell_0$}
         \label{fig:pred_vs_actual_2}
     \end{subfigure}
     \caption{Model performances on the unseen test set for the parameters $m$, $k$, and $\ell_0$.}
     \label{fig:pred_vs_actuals}
\end{figure*}

Nonetheless, the accuracy and loss curves do not provide insight to the unidentifiability of the model. The deep learning model's performance in estimating the subject-wise gait parameters $(m, k, \ell_0)$ on the unseen synthetic test set is presented in Fig.~\ref{fig:pred_vs_actuals}. Accordingly, the model achieves high accuracy in estimating the equilibrium leg length $\ell_0$ from the observation data $(x, y)$, indicating that this parameter is identifiable. However, for parameters $m$ and $k$, the model's performance is less satisfactory. The predictions for $m$ and $k$ display large variability and, as expected, do not converge to the test values, indicating their unidentifiability. This suggests that these parameters cannot be uniquely learned from the observed data alone, even with a deep learning model and abundant training/validation data. 


\subsection{The two-parameter regression problem}
\begin{figure}[tb]
    \centering
    \includegraphics[trim={0in 0 .5in .7in},clip, width=0.8\columnwidth]{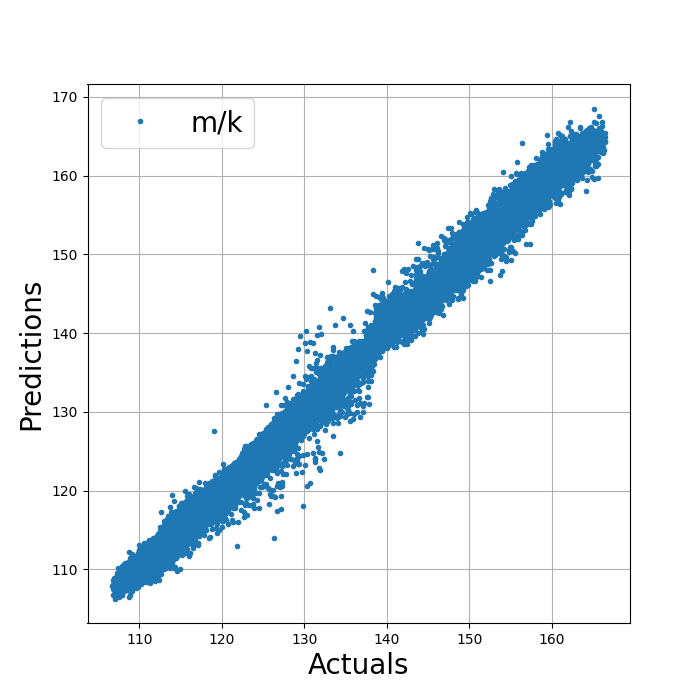}
    \caption{Model performance on the unseen test data in estimating the ratio $\rho = m/k$ (instead of the individual parameters $m$ and $k$).}
    \label{fig:pred_vs_actual_m_div_k}
\end{figure}

To explore the identifiability further, we consider the second scenario shown in Fig.~\ref{fig:2_params_model}, where we seek to estimate $\ell_0$ and the ratio $\rho = m/k$, which combines the unidentifiable parameters $m$ and $k$ into a single parameter. As shown in Fig.~\ref{fig:pred_vs_actual_m_div_k}, the model successfully estimates the ratio $\rho$ with high accuracy, highlighting the model's capability to identify this parameter despite the unidentifiability of $m$ and $k$ individually.

\section{Discussion}

In this study, we explored the notion of identifiability in the context of parameter estimation from motion sensor data using a deep learning model, based on the well-known bipedal-spring mass human walk dynamics model proposed by Geyer et al. \cite{geyer2006compliant}. The results demonstrated that while certain parameters were identifiable from the observation data, others were unidentifiable, indicating that they cannot be uniquely determined solely from the available data. Importantly, the unidentifiability of all or certain parameters does not indicate any flaw in the data model or learning algorithm; rather, it is an intrinsic limitation of the experimental setup, requiring a reevaluation of the data collection and experimental scenario.

The concept of unidentifiability is not unique to our study; various domains encounter similar challenges. For instance, subject localization from two-dimensional images, estimating stochastic parameter priors from aggregated parameter sets, and inferring deep hidden layer parameters from a limited number of observations are all examples of potentially unidentifiable problems. Addressing these issues necessitates revisiting data models and machine/deep learning model architectures, and the development of new data-driven approaches, including algorithms for data fusion and parameter estimation.

To address the challenges posed by unidentifiability in machine learning and deep learning models, several key areas warrant further attention and future studies:

\subsection{Advanced Data Models} Developing more expressive and comprehensive data models that encapsulate the underlying complexities of real-world systems can enhance identifiability. Incorporating domain knowledge and physical constraints into the data models can lead to more informative parameter estimation.
    
\subsection{Multimodality} Depending on the context, integrating multimodal and diverse data sources, such as sensor data, images, videos, and textual data, can provide complementary information and improve the identifiability of model parameters. Fusion algorithms and techniques need to be explored to leverage the full potential of multimodal data assimilation.
    
\subsection{Model-based machine learning} By integrating physics informed (model-based) machine learning with real-world data and leveraging theoretical insights on identifiable/unidentifiable models and systems, we can effectively address unidentifiability challenges. This combination of theoretically supported models and data-driven approaches also enhances model reliability and generalizability. We should emphasize that models are not necessarily accurate or unique, and multiple models, with different levels of abstraction can be considered for a single problem\footnote{For example, the compliant leg model case study that was investigated in this work is only one of the many models that exist for human walking and running modeling.}. Moreover, models are imperfect and are not necessarily fully descriptive of the real world data. Therefore, an open research question is what can be stated regarding the identifiability of systems, for which multiple models coexist.
    
\subsection{Identifiability metrics} Overall, identifiability is an understudied, yet emerging, domain in machine learning and deep learning. Developing metrics to quantitatively assess model identifiability can aid in predicting and diagnosing unidentifiable models. These metrics can guide the selection of appropriate model architectures and data collection strategies to enhance parameter estimation. Considering the interdisciplinary nature of the notion of identifiability and its challenges, collaborations between experts in machine learning, control theory, statistics, and application-specific context knowledge can facilitate a more comprehensive investigation of the topic. Recent research on incorporating the notion of identifiability in machine learning and deep learning is promissing \cite{Roeder2020,Kivva2022}.

\section{Conclusion}
In conclusion, our study provides insights into identifiability in the context of parameter estimation from motion sensor data using deep learning models. We showed that certain parameters can be uniquely determined from the observation data, while others remain unidentifiable --- an intrinsic limitation of the experimental setup. Identifiability extends beyond this case study, impacting diverse domains in machine learning applications. Addressing unidentifiability requires novel data-driven approaches, data source integration, and interdisciplinary collaboration. Future research should prioritize multimodal data integration, model-based machine learning, and the development of identifiability metrics. Collaborative efforts across disciplines will also enable a comprehensive investigation of the notion of identifiability, leading to more informed models and applications. 

\section*{Acknowledgement}
I would like to thank Ms. Taniel Winner for bringing to my attention the compliant leg model by Geyer et al. (2006) in the BMI-532 course on ``Model-based Machine Learning'' and the fruitful conversations with the students, which resulted in a lecture on identifiability of physics-informed data models and the case-study investigated in this research. 
\bibliographystyle{IEEEtran}
\bibliography{References}
\balance

\end{document}